# Fading of collective attention shapes the evolution of linguistic variants

Diego E Shalom[1], Mariano Sigman[2], Gabriel Mindlin[1] and Marcos A Trevisan[1]

[1] Department of Physics, University of Buenos Aires and Physics Institute of Buenos Aires (IFIBA) CONICET, Argentina, [2] Neuroscience Laboratory, Torcuato Di Tella University, CONICET, Argentina

marcos@df.uba.ar

**Language change involves the competition between alternative linguistic forms (1). The spontaneous evolution of these forms typically results in monotonic growths or decays (2, 3) like in winner-take-all attractor behaviors. In the case of the Spanish past subjunctive, the spontaneous evolution of its two competing forms (ended in –ra and –se) was perturbed by the appearance of the Royal Spanish Academy in 1713, which enforced the spelling of both forms as perfectly interchangeable variants (4), at a moment in which the -ra form was dominant (5). Time series extracted from a massive corpus of books (6) reveal that this regulation in fact produced a transient renewed interest for the old form –se which, once faded, left the –ra again as the dominant form up to the present day. We show that time series are successfully explained by a two-dimensional linear model that integrates an imitative and a novelty component. The model reveals that the temporal scale over which collective attention fades is in inverse proportion to the verb frequency. The integration of the two basic mechanisms of imitation and attention to novelty allows to understand diverse competing objects, with lifetimes that range from hours for memes and news (7, 8) to decades for verbs, suggesting the existence of a general mechanism underlying cultural evolution.**

Alternative linguistic forms such as synonyms (garbage-rubbish), spelling differences (behavior-behaviour) or past-tense regularization (spilt-spilled) are in constant tension and present a unique case to study the dynamics of cultural transitions (2, 9). This rivalry between competing cultural expressions occurs in a time-scale of centuries or decades, but similar phenomena can be observed also in the time scale of hours for memes (8) and days in news (7).

Cultural transitions have been modeled with a variety of approaches going from formal and normative approaches (8, 10) to some that explicitly incorporate the cognitive mechanisms that may produce them (2, 7, 11, 12). Distinct versions of the cognitive approach have proposed three mechanisms that account the fading of certain cultural forms, being replaced by alternative versions which dominate the scene: 1) limited attention, 2) imitation, and 3) preference for novelty or adaptation (7).

Recently, a specific mathematical model (2) was proposed to explore transitions in language which included the Spanish subjunctive. This presents a quite privileged window to investigate cultural rivalry because it comes in two virtually perfectly exchangeable variants (13–16) (ended in –ra or –se, as in *cantara* and *cantase*, from the verb cantar, to sing), which allows to isolate and model specifically the selection forces that bias the speakers towards either alternative. Starting approximately in 1800, written word usage reveals a monotonic decay of the form –se, which is replaced by the form –ra that dominates the Spanish subjunctive to this day. The model by Amato and collaborators (2) can explain these dynamics, since it predicts monotonic trajectories of competing populations. However, as one goes back in the history of the subjunctive, as we detail below, one finds a non-monotonic dependence with a transition in which the form –se increases its popularity, which is then followed indeed by a slow and progressive decay in which the form –ra takes over again. This more interesting behavior resembles the dynamics of a damped oscillator in critical regime, suggesting that the dynamics of this transition is governed by a two-dimensional linear process.

The aim of the work described in this letter is two-fold: 1) to understand how non-monotonic cultural transitions can be accounted for by models which include imitation and novelty as used in different domains and temporal scales of culture. 2) To understand, from a normative perspective, how novelty, which can be modeled as a dynamic forcing of one of the competing forms, leads to a minimal form of competing model equivalent to critically damped oscillators.

We begin with a brief history of the Spanish subjunctive, which presents several outstanding aspects that shaped its evolution over the last six centuries. The subjunctive is a grammatical mood that expresses states of unreality such as wish, possibility or action that have not yet occurred (e.g. 'Wish you were here'); this mood is often contrasted with the indicative, used mainly to point that something is a statement of fact (e.g. 'You are here') (14). Notably, -ra is the only verbal form in the history of Spanish that changed its mood, from its Latin indicative value (which was preserved in Old Spanish) to its modern subjunctive value. Ralph Penny says that, from the fourteenth century on, -ra "began to be used as an imperfect subjunctive, coming into competition with -se and eventually ousting this form in many varieties of Spanish" (5). A long process mediated the ascription of these forms to the subjunctive, and it was not until the seventeenth century that they began to occupy contexts reserved to the –-se form, first in final subordinate sentences and then in the rest of the contexts (17).

One century later, in 1713, the Royal Spanish Academy (RAE) was conceived as the official institution responsible for overseeing the Spanish language and promote linguistic unity within and across Spanish-speaking territories (2, 4). The Academy produced a global standardization process that enforced the official spelling of a number of linguistic forms, including –ra and –se as completely interchangeable variants of the past subjunctive in any context, without change of meaning (5, 17, 18).

Our conjecture is that, in a context where the form –se was declining (5), this would produce a percolating effect of novelty which should lead, as in a critically damped oscillator, to a sharp transition of the –se form followed by a reestablishment to the preferred –ra form once the novelty effect lost its traction.

To explore the dynamics of the equivalent forms after the institutional regulation, we collected tokens of Spanish verbs from the two main Spanish corpora, the Spanish Google books (6) and the New Historical Spanish Dictionary (CDH) from the Royal Spanish Academy (19) (see Methods). Figure 1a shows the frequency of the first 100 verbs in subjunctive mood, computed in the period 1700-2000. As expected, verbs follow the Zipf´s law (20).

From all collected tokens, we computed the number of each subjunctive variant $N_{ra}$ and $N_{se}$ on a yearly basis. In Figure 1b we show traces of the relative frequency for –se, $s = N_{se}/(N_{ra} + N_{se})$ for verbs of different frequency in the list. The traces show that the –se form systematically increased its popularity during the eighteen century, which is also supported by analyses on small private corpora (13). Then, after attaining a maximum value around 1800, the trend reversed and the form -se decreased monotonically, making –ra the dominant form observed at the present time across the Spanish speaking countries (13).

As we go back in time, the number of books per year diminishes and the collected tokens become increasingly sparse. We therefore bounded our analyses to the period beginning at year 1750 (vertical line) for which subjunctive instances for all verbs are greater than 1000 tokens per year in the Google Books corpus (Figure S1). The amount of tokens per year depends in turn of each specific verb. We required the existence of subjunctive instances in each 5-year window within the period 1750-2000, which holds for the top 40 most frequent verbs (see Methods).

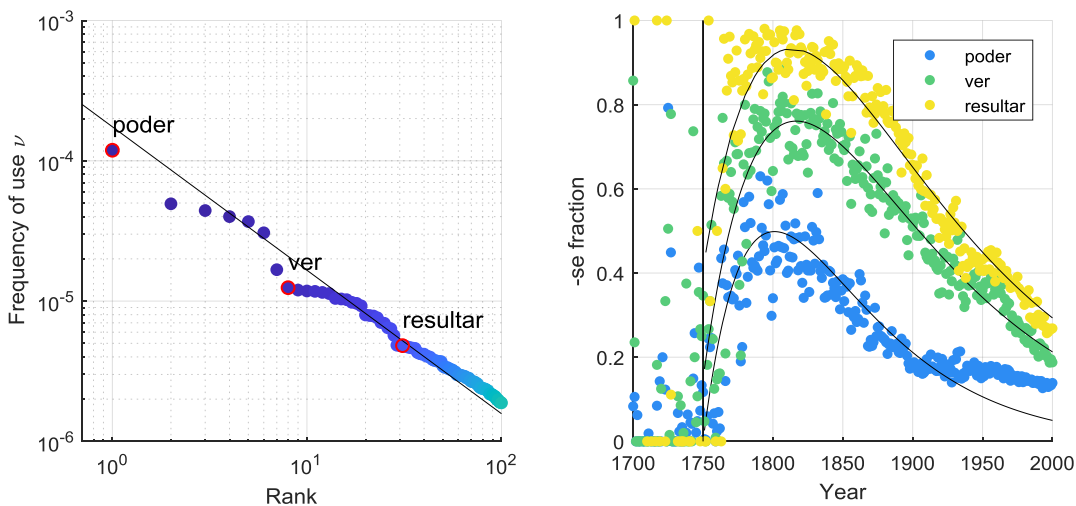

**Figure 1 | Evolution of the fraction -se in the period 1750-2000. a**, we computed the frequency of subjunctive tokens (normalized to the size of the corpus) for the 100 most used Spanish verbs, which follow the Zipf's law. **b,** time traces of the –se fraction of poder (to be able), ver (to see) and resultar (to result), representatives of high, medium and low frequency of use (Spanish corpus of

Google Books). Time traces show an overshoot circa yr. 1800. The vertical line at yr. 1750 indicates the cutoff point after which the density of forms –ra and –se is enough for data analysis.

Recently, a discrete model for the competition of linguistic variants under different enforcements was presented (2). Notably, control on a few parameters allows fitting the model to experimental data of spontaneous competitive processes, or under the action of formal or informal authorities. In particular, the model was used to fit the evolution of the fractions $s$ and $r$ in the period 1800-2000. According to the model, at each year $t$ writers contribute to the corpus of books in this way: a fraction $c$ of them use the -ra convention; the rest of them (fraction $1 - c$) either follow the institutional enforcement $E$ (fraction $\gamma$) or sample the current distribution (fraction $1 - \gamma$). With these ingredients, the model reads:

$$r_{t+1} = (1-c)[\gamma E_r + (1-\gamma)r_t] + c$$
$$s_{t+1} = (1-c)[\gamma E_s + (1-\gamma)s_t]$$

The dominance of –ra is guaranteed by the asymmetry introduced by the parameter $c$ in the first equation. The fractions $E_r$ and $E_s$ represent the institutional enforcement that bias writers towards either convention. Setting $E_r = E_s = 1/2$ to account for the equivalence of both forms, the authors of the work successfully fitted the time series of the variants in the period 1800-2000, when –se decreases monotonically (2).

We capitalized on this to model the evolution of the fractions throughout all the temporal window for which there is sufficient data (from the early eighteen century to these days) which displays, as mentioned above, a non-monotonic progression from the early eighteenth century on. We first assumed no intrinsic assymmetry between conventions ($c = 0$). Taking the continuum limit ($x_{t+1} - x_t \to \dot{x}$) and defining $\tau = 1/\gamma$, the equation for –se reads

$$\tau \dot{s} = -s + E_s(t), \qquad (1)$$

with an equivalent equation for the fraction –ra ($r$). Here, the function $E_s(t)$ does not represent the bias produced by the regulation itself, but the effect of a renewed attraction towards the form –se exerted by the Academy enforcement, in a context where the spontaneous competition between both forms had already established a preference for the variant –ra. Setting

$$E_s(t) = a\tau\, e^{-t/\tau} \qquad (2)$$

we assumed that attention was attracted towards the ousted form –se, which faded exponentially among the population. This choice implies the eventual dominance of –ra, $E_s(\infty) = 0$. The resulting equation is solved by the functions

$$s(t) = (at + s_0)\, e^{-t/\tau} \qquad (3)$$

that overshoot the stimulus $E_s(t)$ and then follow the exponential decay, a picture that qualitatively describes the behavior of the time series shown in Figure 1b, where –se gains popularity before decaying (Figure 1b and Supplementary Figure S3). To quantify this, we associated each verb with the pair of parameters $(s_0, \tau)$ in eq. (3). The value of $s_0$ represents the –se fraction at the beginning of the period (we set $t = 0$ at yr. 1750) and $\tau$ is the lifetime of the exponential decay $E_s(t)$. A single $a$ value was calculated for the complete set of verbs, $a = 0.027 \pm 0.004$ (see Methods).

Figures 1b and S3 show the time traces and fitted curves within the period 1750-2000. We verified that the values obtained for $s_0$ are positively correlated with the values of the –se fraction at year 1750, computed from the independent CDH corpus (Figure 2a). More interestingly, Figure 2b shows that relaxation times $\tau$ are inversely related to the frequency of use $\nu$ of the verbs, following the power law $\tau = \nu^{-\beta}$ ($\beta \sim 0.14$).

In this description, $E_s(t)$ is interpreted as the fraction of writers driven by attention to the variant -se, and consequently $E_s(t) = a\tau e^{-t/\tau} \leq a\tau \leq 1$. As expected by the statistical interpretation of the model, this holds for low relaxation times ($\tau \leq 1/a \sim 43$ years, see Figure 2b) that correspond to highly popular verbs. However, the model successfully fits experimental traces for all verbs, even those of much lower frequencies than the required by the statistical interpretation. This invites to consider the system of equations 1 and 2 as a more abstract model for the variants, which can be rewritten as the autonomous 2-dimensional linear system

$$\begin{pmatrix} \dot{s} \\ \dot{e}_s \end{pmatrix} = \begin{pmatrix} -1/\tau & 1 \\ 0 & -1/\tau \end{pmatrix} \begin{pmatrix} s \\ e_s \end{pmatrix}$$

where $e_s = E_s/\tau$. For this system, trajectories in phase space $(s, e_s)$ approach the origin without oscillations and faster than any other linear system. This borderline case describes many natural phenomena, as cascades of chemical reactions (21) and critically damped oscillators (22). From this abstract perspective, the onset of the regulation acts as a perturbation $e_s(0) = a$ that kicks the system away from the equilibrium $(0,0)$, to which it then relaxes with a timescale $\tau$ inversely proportional to the frequency $\nu$ of each verb.

The competition of subjunctive forms seems to be dominated by a general preference for the form –ra across the Spanish speaking countries (23), which is indeed manifested in the fact that the institutional enforcement produced just a local perturbation in the trend of increasing dominance of –ra. Forces that drive selection to one of the alternatives in competition have been explored, and include social and cognitive factors, as well as phonological and biophysical factors (3, 24). Although the forces underlying the preference for –ra rest unknown, the dynamics that drive speakers towards the preferred form are modulated by collective behaviors such as the ones analyzed in the present work.

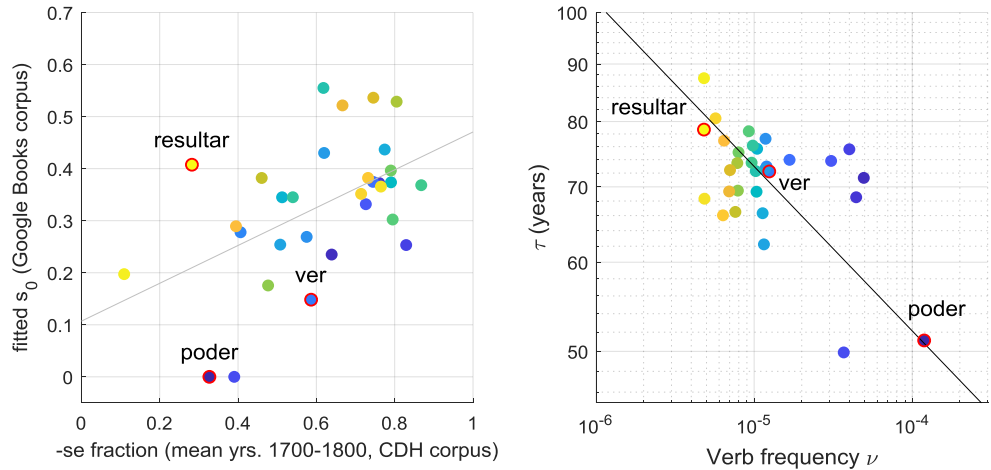

**Figure 2 | Relaxation times decrease with verb popularity. a,** the values of $s_0$ in equation 2 represent the fraction –se at year 1750, fitted with our model using data from the Spanish corpus of Google Books in the period 1750-2000. These values are positively correlated with the mean value of –se in the period 1700-1800 from the independent CDH corpus (r=0.52, b=0.41, p<0.002). The colormap covers the range of verb frequencies, from high (blue) to low (yellow) values. **b,** the most used verbs are less sensitive to novelty effects, evidenced by the inverse relationship between the relaxation timescale $\tau$ and frequency of use $\nu$ (Deming corrected linear regression, r=0.50, p<0.003).

Here we capitalized on the access to massive digital corpora (19, 25) and previous models (2, 7, 10) to advance a simple dynamical system that fits experimental traces of imperfect subjunctive variants collected from written texts. The model integrates two basic forms of collective behavior: an imitative component (writers contribute to the corpus by sampling the current distributions of the variants) and a novelty component (writers temporarily pay attention to the declining form).

Our main findings are that 1. the model successfully fits experimental time traces starting at year 1750, close to the external regulation that stated the equivalence of the variants. This regulation acted as a perturbation that unfolded the underlying dynamics of the problem, compatible with a linear system in critical regime. 2) adjusting the dynamics for each verb revealed that the relaxation timescales are in inverse proportion to the verb frequency, which means that popular verbs are less prone to change by novelty. This contributes to form a picture of verbal change, beyond the observed resistance of the most popular verbs to regularization (11).

Imitation and preference to novelty have been also suggested as necessary components for modelling data obtained from the competition of news in the web (7). The integration of these two basic behaviors allows then to understand diverse competing objects, with lifetimes that range from some hours or days for memes (8) and news (7) to decades for verbs. This suggests a general mechanism underlying cultural evolution, and provides us with a general framework to study cultural phenomena.

**Methods summary**

We extracted the list of 889 Spanish verbs with more than one instance of imperfect subjunctive from the LexEsp corpus (26). For each verb, we built the list of all conjugations in both imperfect subjunctive forms: –ra (cantara, cantaras, cantáramos, cantárais, cantaran) and –se (cantase, cantases, cantásemos, cantáseis, cantasen). These tokens were then collected from the Google Books (25) and CDH (19) corpora. From all collected tokens, we selected those corresponding to verbs that occurred at least once in each 5-years window from 1750 on (see supplementary Figure S1). This condition was met for the top 40 most used Spanish verbs in the Google books corpus. The frequency of use of each verb was computed as the ratio between all the conjugations for the simple past subjunctive collected in the corpus and the size of the corpus ($8.4 \ 10^{10}$ tokens).

In the absense of regulation, many orthographic forms coexisted in Old Spanish. For instance, the verb haber (to have) included the following –se orthographic variants for the imperfect subjunctive: oviese, oviesse, hobiese and hubiese. We verified that the archaic forms were absent after 1700 (see Supplementary Figure S2) and therefore only the modern past subjunctive forms were collected and analyzed.

Four verbs were explicitly excluded from our analyses. The verbs ser (to be) and haber (to have) serve a different function, as auxiliars for the compound tenses in Spanish. Also, the verb ir (to go) shares the same imperfect past subjunctive with the verb ser (to be), i.e. fuera/fuese. For these reasons, those cases were a priori excluded from our analyses. The time trace of the verb deber (should) was not fitted by the model and was also excluded from the analysis.

Parameters $a, \tau$ and $s_0$ of equation (3) were fitted to the experimental time traces using a standard least squares method (Figure S3). The pairs $(s_0, \tau)$ for each verb in the set were calculated as a function of $a$, whose value was then set as the one that minimizes the sum of errors for all verbs. The range of $a = 0.027 \pm 0.004$ was determined by bootstraping (100 repetitions of 2 randomly chosen verbs).

**Acknowledgements**

Marcos Trevisan thanks Damián Decuzzi and María Giannoni for beautiful conversations. The research in this work was partially funded by the Concejo Nacional de Investigaciones Científicas y Técnicas (CONICET), the University of Buenos Aires (UBA) and NIH through R01-DC-012859.

**Supplementary Information**

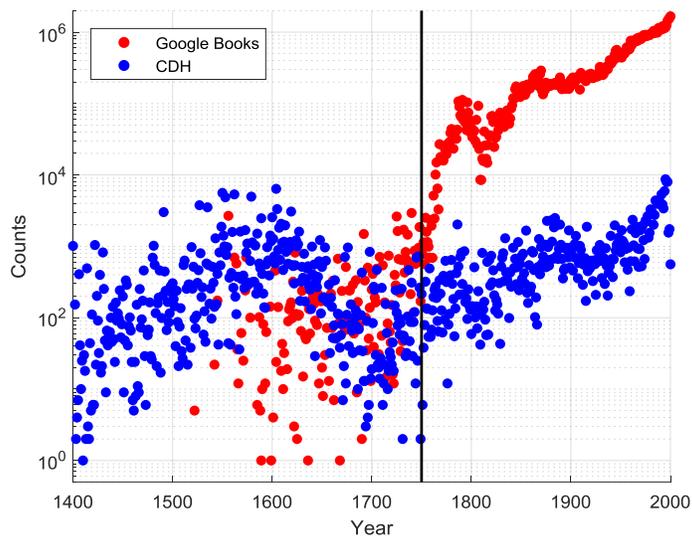

**Figure S1 | Past subjunctive tokens for the 50 most used Spanish verbs in the Google Books (red) and CDH (blue) corpora.** Counts of Spanish past subjunctive verbal forms in the two corpora.

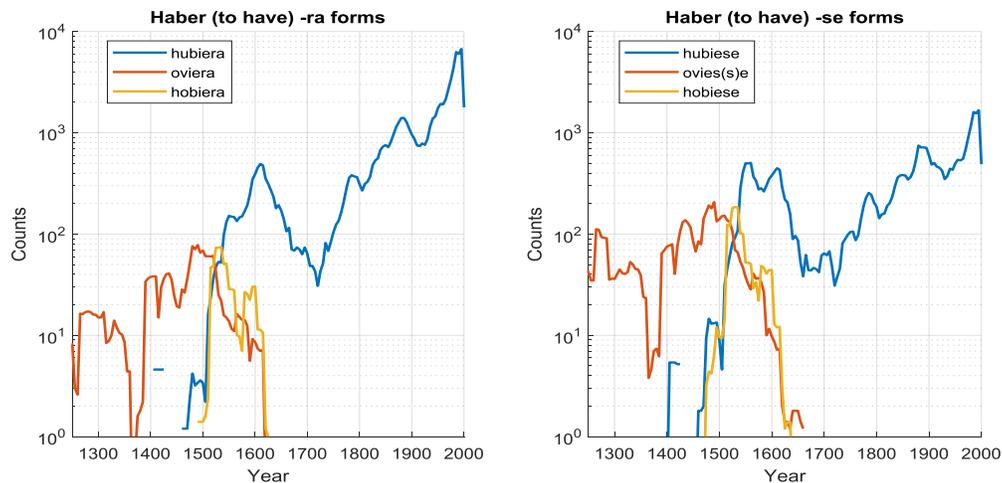

**Figure S2 | Modern verbal forms for –ra and –se variants became standardized at the early eighteen century (CDH corpus).** Blue lines represent appearances of the modern past subjunctive of the verb haber (to have) ended in –ra (left panel) and in –se (right panel). Only these forms (hubiera and hubiese) are present in the period analyzed here, from yr. 1700 to 2000. In absence of regulation, different orthographic variants coexisted (as shown for the forms oviera and hobiera, hobiese, oviese, oviesse and hubiese in the period circa 1500-1700). Counts correspond to 5 years-windows.

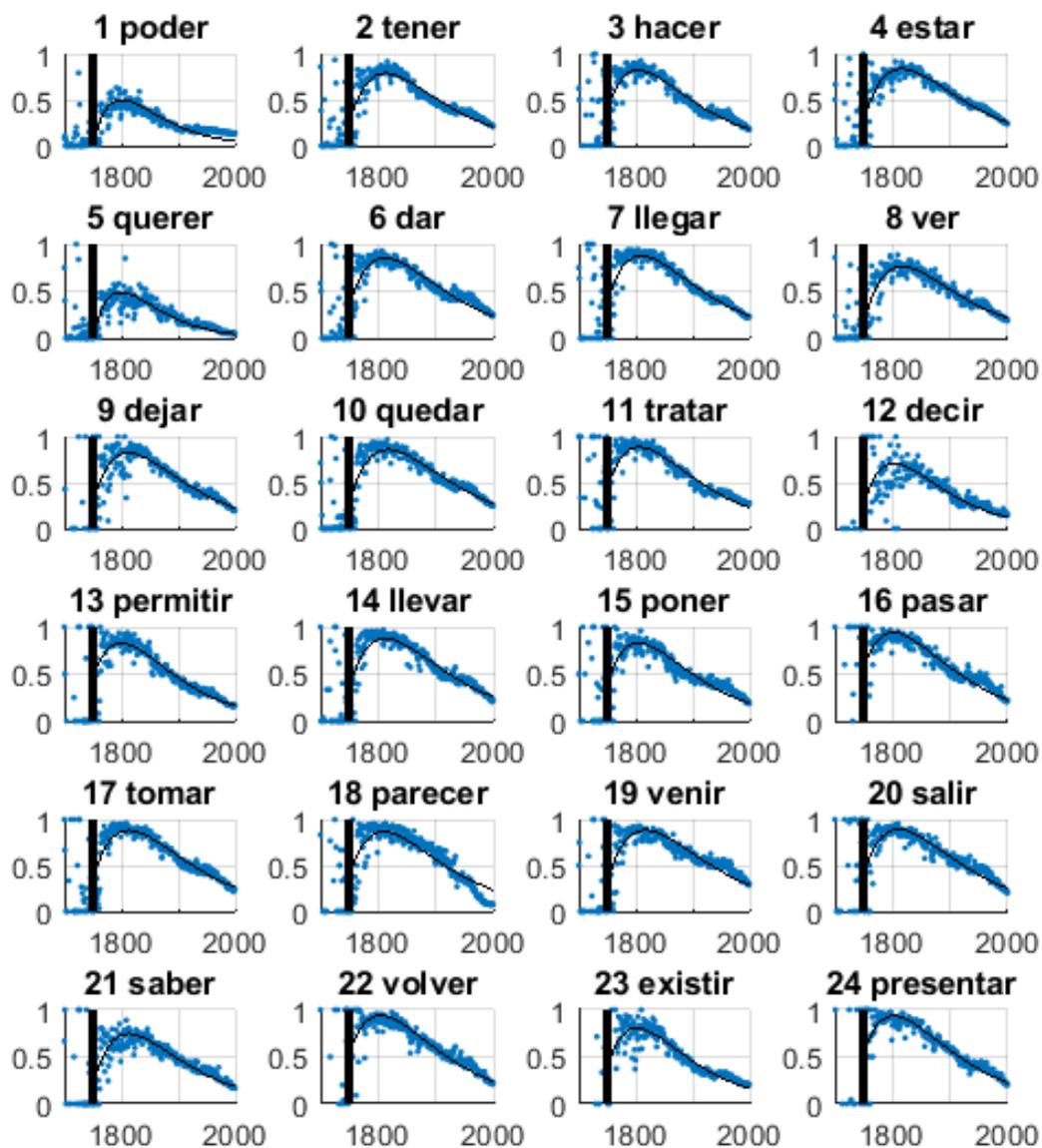

**Figure S3 | -se fraction in the period 1700-2000 and model fitting.** Data from Google Books corpus for the first 48 most used verbs in Spanish.

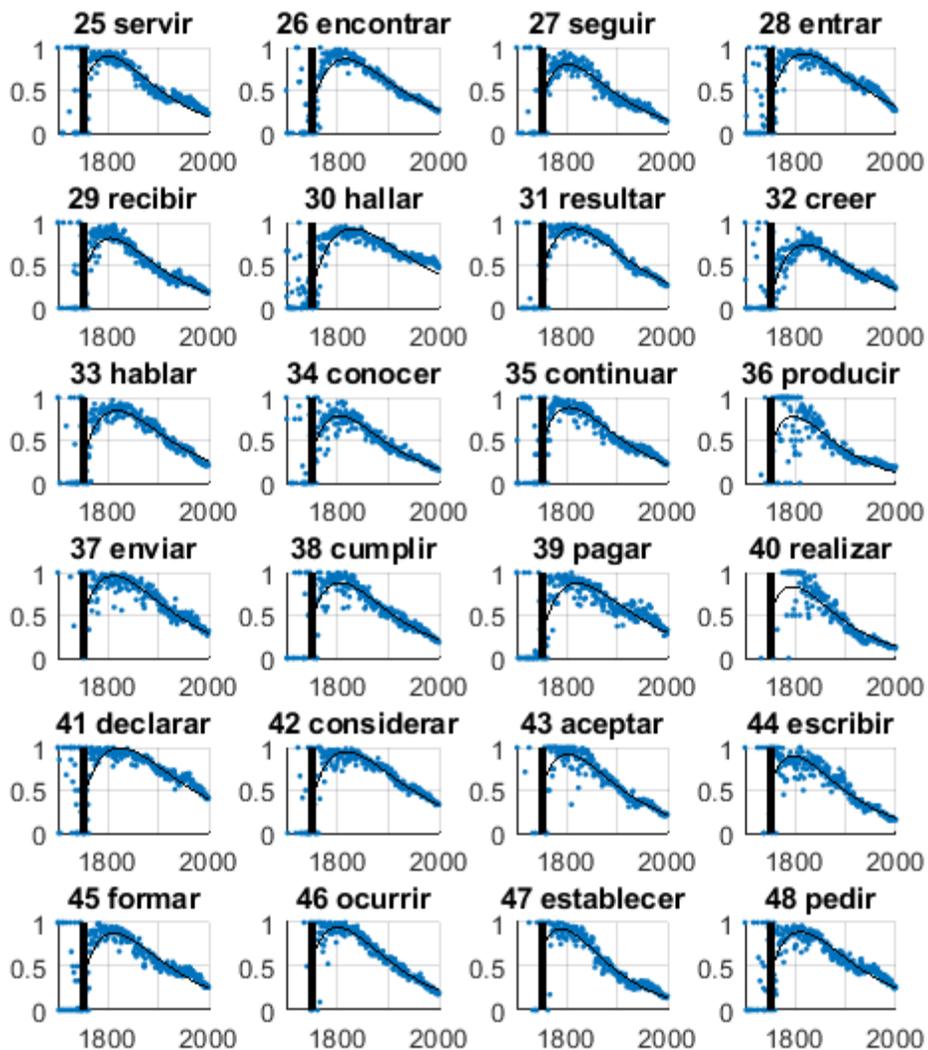

**Figure S3 (cont.) | -se fraction in the period 1700-2000 and model fitting.** Data from Google Books corpus for the first 48 most used verbs in Spanish.